\documentclass[pdflatex,sn-mathphys-num]{sn-jnl}

\usepackage{graphicx}%
\usepackage{multirow}%
\usepackage{amsmath,amssymb,amsfonts}%
\usepackage{amsthm}%
\usepackage{mathrsfs}%
\usepackage[title]{appendix}%
\usepackage{xcolor}%
\usepackage{textcomp}%
\usepackage{manyfoot}%
\usepackage{booktabs}%
\usepackage{algorithm}%
\usepackage{algorithmicx}%
\usepackage{algpseudocode}%
\usepackage{listings}%

\usepackage{array}
\usepackage{stfloats}
\usepackage{url}
\usepackage{verbatim}
\usepackage{paralist}
\usepackage{placeins}

\theoremstyle{thmstyleone}%
\theoremstyle{thmstyletwo}%

\theoremstyle{thmstylethree}%

\begin{document}

\title[Article Title]{SCoPE: Shift-Aware Speaker-Conditioned Priors for Emotion Recognition in Conversations}


\author*[1]{\fnm{Burak Can} \sur{Kaplan}}\email{burak.can.kaplan@uni-hamburg.de}

\author[1]{\fnm{Stefan} \sur{Wermter}}\email{stefan.wermter@uni-hamburg.de}

\affil*[1]{\orgdiv{Department of Informatics}, \orgname{University of Hamburg}, \orgaddress{\street{Bundesstraße 56b}, \postcode{20146}, \city{Hamburg}, \country{Germany}}}


\abstract{In conversations, human emotions are transient; however, they tend to persist across multiple utterances. For example, we rarely switch instantly between contrasting emotions such as happiness and anger. Instead, emotions tend to evolve smoothly, and these patterns are often speaker-specific. Some people might escalate, while others gradually cool down over time. Furthermore, when emotions change during a conversation, they are often driven by contextual factors, such as newly received information or unexpected events. Even though progress has been made in Emotion Recognition in Conversations (ERC), most existing approaches still rely heavily on overt evidence and do not sufficiently model these non-apparent factors. Especially in multimodal settings, this makes these models fragile when the signals are noisy (e.g., occluded faces, slang expressions, or microphone noise). To address these limitations, we introduce Speaker-Conditioned Priors over Emotions (SCoPE). SCoPE is a light weight module that utilizes the emotional history of each speaker and explicitly models their priors for use in subsequent emotion classification. Second, we incorporate emotion shift prediction, a well-established concept in ERC, to guide the model in balancing the priors from SCoPE and multimodal evidence. Finally, we propose a shift-aware fusion mechanism that performs precision-weighted logit integration between multimodal evidence and the speaker prior, forming a Bayesian-inspired product-of-experts formulation. This dynamic fusion allows the model to rely on historical priors when emotions persist and to prioritize multimodal evidence when shifts are likely. Experimental results show our model achieves superior performance over recent state-of-the-art models on the IEMOCAP dataset in multimodal settings.}




\maketitle

\section{Introduction}
\label{sec:intro}
Emotion Recognition in Conversations is a task of understanding and modeling the persons' emotions in interactions\cite{8764449, poria-etal-2019-meld}. ERC models use dialogues as input and examine the sequential utterances for each speaker. The dialogues can be in turn-taking, or in random turns, which is more natural in multi party settings. ERC has a great potential to enable emotionally intelligent Human-Robot Interaction (HRI) systems~\cite{picard2000affective, BREAZEAL2003119, affectdetection, affectreview}. Unlike traditional Emotion Recognition, ERC does not aim to learn isolated per-utterance representations~\cite{zhong-etal-2019-knowledge, ghosal-etal-2019-dialoguegcn, ghosal-etal-2020-cosmic}. Instead, it learns the contextual information and the dynamic nature of emotions, which is much closer to reality.

ERC comes with multiple challenges, arising from the nature of everyday dialogues~\cite{Majumder_Poria_Hazarika_Mihalcea_Gelbukh_Cambria_2019, hu-etal-2021-dialoguecrn}. First, it requires strong reasoning capabilities over multiple speakers, because speaker turns often include interconnected emotional trajectories. ERC in real time is more than just classification since it is an inference over multiple temporal states. Second, ERC also requires reasoning ability to understand what the speakers truly express. Since emotions are not only expressed by the text, there are many clues in other modalities as well. However, handling multiple modalities also brings multiple challenges~\cite{misa, multimodallearn}. It introduces noise, missing or misleading signals, and sometimes even contradictions. For example, it is a big challenge to model sarcasm, in which the facial image might disagree with the context. 

\begin{figure}[t]
\centering
\includegraphics[
    width=0.4\textwidth,
    trim=0 0 0 18.7cm,
    clip
]{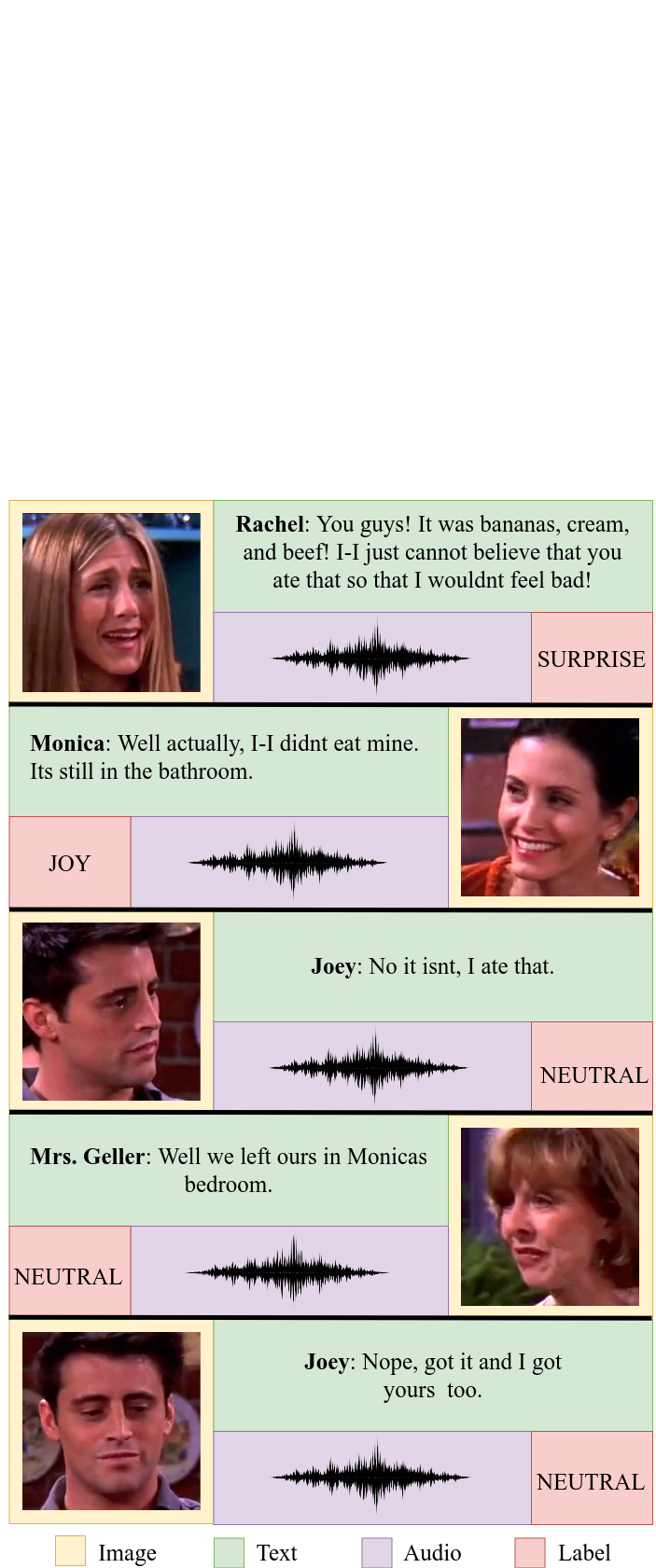}

\caption{\centering Multimodal ERC data example, from MELD~\cite{poria-etal-2019-meld} dataset.}
\label{fig:sample}
\end{figure}

Figure~\ref{fig:sample} shows an example of multimodal ERC data. In the ERC data, we have observable cues such as the content, facial expressions, prosody of the voice, etc. These cues provide explicit evidence to the model. However, emotions are also shaped by latent factors such as: 
\begin{itemize}
    \item Personality of the speaker
    \item Mood of the speaker
    \item Speaker-specific Tendencies
\end{itemize}
These factors are often not directly observable but still have a strong effect on emotional evolution~\cite{kuppens2010emotional, Koval2021, fleeson2001toward, robinson2002belief}. Current ERC models still mainly focus on multimodal evidence and do not model those signals explicitly. However, a better ERC model should also take these latent factors into consideration to understand the speaker's emotion. 

In conversations, emotions are temporary states; however, they still tend to persist across multiple turns~\cite{KOVAL20243, waughetal, kuppens2010emotional, affectivedynamics}. In other words, humans do not shift instantly between contrasting emotions, such as from happy to anger. Instead, the transitions between emotionally distant states are sparse and mostly gradual~\cite{Scherer01112009, bagnara2025feelings, shao2026seven}. Also, the evolution of emotions is different for every person with the effect of the aforementioned latent factors. For example, some people might escalate their feeling whereas other people can cool down over time. Due to this gradual evolving nature of emotions, past emotional states provide a strong predictive signal for future states that can be used for ERC.

Emotional persistence often happens for multiple turns, but emotions do not have to stay the same through an entire dialogue. Since they may linger for some turns and a shift may happen. These shifts are often triggered by factors such as receiving new information, social feedback, unexpected events, or changes in tone or interaction dynamics~\cite{lazarus1991emotion, scherer2001appraisal}. So when an emotional shift happens, it is critical to check the current multimodal evidence to reveal the underlying factors. Therefore, ERC models must reason about when to trust the emotion history and when to override it.

Many of the current ERC models still largely focus on per-utterance multimodal features, which makes temporal modeling of the emotions often implicit or shallow. A system that only looks at per-utterance multimodal evidence can be fragile when the signal is noisy, such as occluded faces, slangy text, or microphone noise. On the other hand, recent approaches show that tracking emotion shifts is important and has a positive effect on ERC classification in terms of accuracy~\cite{ghosal-etal-2020-cosmic, wang-mine-2023-multi}. However, these approaches also treat emotion shifts as labels, not as control signals. Moreover, there is very limited integration of psychological priors into ERC architectures, which leaves potential for improvement for this field.

In conversations, people do not interpret emotions as isolated observations. Instead, they implicitly have a belief about the emotional state of other speakers as the interaction continues~\cite{friston2010free, barrett}. This belief acts as a prior: ”Given how this person felt so far and who is speaking now, what is the most likely state next?” Importantly, this belief is not always constant. When an emotion shift is unlikely, humans tend to rely more on their existing belief about a speaker’s emotional state, whereas in moments of potential change, current contextual and multimodal evidence becomes more influential.~\cite{Wilkinson2019, kube} So effective ERC requires not only emotional persistence, but also reasoning about when this persistence should be changed. Motivated by this intuition, we formulate the following research questions: (1) how accurately can emotion shifts be predicted from conversational context, (2) how can emotional state persistence be modeled explicitly in a speaker-conditioned manner, and (3) how can shift prediction be used to guide temporal emotion reasoning in conversational settings?

To tackle these research questions, first we introduce a dual-head ERC framework that has an emotion classification head and an emotion shift prediction head. Second, we introduce Speaker-Conditioned Priors over Emotions, shortly SCoPE. SCoPE is a GRU-based module which is designed to simulate implicit prior states to complement our dual-headed framework. SCoPE is (1) lightweight, (2) autoregressive, and (3) speaker-aware. Finally, we introduce a shift-aware fusion to support the model balance between speaker priors and multimodal evidence. Using this shift-aware fusion, the history is emphasized when the model detects that an emotion shift is unlikely. On the other hand, multimodal evidence is emphasized more when the shift is likely. Our whole model remains lightweight, fast, and end-to-end trainable.

The contributions of this paper are as follows:
\begin{itemize}
    \item We introduce SCoPE as a novel speaker-conditioned temporal prior generator for ERC.
    \item We use emotion shift as a control signal rather than an auxiliary task.
    \item We propose a shift aware fusion mechanism that dynamically balances prior and evidence for better ERC classifications.
    \item We demonstrate that our framework delivers consistent performance improvements against its baseline and gives superior performance on popular ERC benchmarks. 
\end{itemize}

This paper is structured as follows: Section~\ref{sec:related} provides prior work done on ERC in AI and psychology to give a grounding for our motivations. Section~\ref{sec:method} explains the details about our baseline, methodology and modules. Section~\ref{sec:experiments} provides detailed information about the datasets and experimental setup. Our results will be discussed in Section~\ref{sec:results}. Finally, the overall conclusion will be provided in Section~\ref{sec:conc}, as well as some ideas about the future work.

\section{Related Work}
\label{sec:related}
\subsection{ERC Overview and Multimodality}

ERC started to get more attention as a task in 2019 \cite{8764449}, and early ERC models classify emotions at context-aware utterance level~\cite{poria-etal-2017-context}. Architecture-wise, previous ERC models can be grouped into RNN-based \cite{hazarika-etal-2018-icon, Majumder_Poria_Hazarika_Mihalcea_Gelbukh_Cambria_2019}, Transformer-based \cite{li-etal-2020-hitrans, Ma_2024} and Graph-based \cite{ghosal-etal-2019-dialoguegcn, Tu_Xie_Liang_Wang_Xu_2024}. With the emergence of Large Language Models (LLMs) recently, ERC classification has also been explored by general LLMs and LLM-based ERC architectures \cite{Tu_Xie_Liang_Wang_Xu_2024, lei2023instructerc, wu2026reinforcement}. However, they are costly to train, and it has been observed that the performance of general models like ChatGPT is still not as strong as the models specialized for multimodal ERC \cite{fu-etal-2025-laerc}. 

Initially, some ERC models focused on text-only classification \cite{lee-lee-2022-compm, song-etal-2022-supervised}. Because visual and audio features can introduce more noise than text, this may lead to a decrease in the model's performance \cite{wu-etal-2025-multimodal}. Also, high quality multimodal ERC data is quite scarce, and building such datasets is costly and challenging. On the other hand, enhancing the text of existing datasets \cite{xue2024bioserc} or generating more text-based ERC datasets is easier, especially due to the availability of LLMs \cite{kaplan2025can}. However, text only is not enough nowadays, especially given that advanced AI assistants and HRI systems are getting more important and getting more attention. Therefore existing multimodal datasets such as MELD \cite{poria-etal-2019-meld} and IEMOCAP \cite{busso2008iemocap} remain valuable for the field and recent research has increasingly focused on multimodal ERC \cite{wu-etal-2025-multimodal}. While these approaches model contextual dependencies, they often focus on extracting discriminative multimodal features rather than explicitly modeling emotion dynamics. 

\subsection{Emotion Shift Detection and Modeling}

Emotion Shift Detection (ESD) is a task of detecting the emotion shift of the same speaker in consecutive utterances. ESD was introduced as an auxiliary task to ERC to explicitly model the transitions in ERC scenarios \cite{GAO2022108861}. It is a proven method to increase the classification performance of the models \cite{chen2022emotion, GAO2022108861}. Recently, prior approaches used ESD to explicitly predict emotion transitions or shifts to be used for ERC \cite{wang-mine-2023-multi, li2024cfn, yang2025graph}. Emotion shift has been used as:  \begin{itemize}
    \item Auxiliary task \cite{GAO2022108861, li2024cfn, wang-mine-2023-multi},
    \item Regularization constraint \cite{chen2022emotion, yang2025graph},
    \item Evaluation metric (Emotion Flip) \cite{kumar-etal-2024-semeval, kumar2022discovering},
\end{itemize}
However, emotion shift could be exploited much more beneficially. Many approaches use shift as an auxiliary supervision signal during training, but do not integrate it into inference as a dynamic controller of temporal priors. In contrast, we use emotion shift prediction as a control signal that dynamically regulates the influence of speaker-conditioned temporal priors against multimodal evidence.

\subsection{Temporal Dynamics and Sequential Emotion Modeling}

Modeling temporal dependencies is a core challenge in ERC, as emotional states evolve across utterances rather than appearing independently. Early ERC approaches use recurrent architectures to capture contextual dependencies in dialogue turns \cite{hu-etal-2021-dialoguecrn, Majumder_Poria_Hazarika_Mihalcea_Gelbukh_Cambria_2019}. On the other hand, later approaches use graph-based \cite{ghosal-etal-2019-dialoguegcn, sun-etal-2021-discourse-aware, shen-etal-2021-directed} or transformer-based \cite{kim2021emoberta, Ma_2024} ERC models to extract context information.

Prior studies explore the sequential nature of emotional expressions explicitly \cite{wang-etal-2020-contextualized, liang-etal-2022-page, ghosal-etal-2019-dialoguegcn}. They either model emotion transitions or use the temporal consistency across utterances. For example, recurrent and conditional random field–based approaches are used to smooth emotion predictions to reduce extreme changes in predicted labels \cite{song2022emotionflow, liang-etal-2022-page}. These methods show that using the temporal information often improves the ERC classification performance.

Most of these approaches assume uniform emotion dynamics across speakers and the context. However, temporal dependencies are learned implicitly, from hidden states or attention mechanisms without modeling emotions as latent states with decay. Additionally, the influence of the chat history is often fixed, regardless of the chance of emotion shift happen. As a result, current models may either over-smooth emotional trajectories or react too strongly to local noise. This motivates more adaptive and structured approaches for temporal emotion modeling.

\subsection{Speaker Modeling and Speaker States in Dialogue}

Speaker information plays an important role in ERC, as most ERC scenarios involve multiple speakers \cite{poria-etal-2019-meld, busso2008iemocap}. It is common that ERC models use speaker embeddings to learn speaker-specific behaviors \cite{Majumder_Poria_Hazarika_Mihalcea_Gelbukh_Cambria_2019, saxena-etal-2022-static}. For example, DialogueRNN maintains speaker-dependent latent states and aims to model context speaker specifically \cite{Majumder_Poria_Hazarika_Mihalcea_Gelbukh_Cambria_2019}. Memory-based architectures such as CMN \cite{hazarika-etal-2018-conversational} and ICON \cite{hazarika-etal-2018-icon} explore self- and inter-speaker influences through memory mechanisms.

However, in existing ERC models speaker information is often treated as either a static conditioning factor \cite{saxena-etal-2022-static}, or as part of a learned representation to optimize classification \cite{wang-etal-2024-emotion, Li_Zhu_Mao_Cambria_2023}. These approaches capture who is speaking, but they do not explicitly model how emotions evolve differently for different speakers over time. Moreover, existing approaches often lack mechanisms to learn speaker-conditioned emotional tendencies, such as escalation, gradual cool-down, or persistence \cite{KUPPENS201722, waughetal}. However, these speaker behavior patterns are common in real-world interactions. This motivates the development of models that use speaker identity to guide detection of the temporal evolution of emotions explicitly.

\subsection{Psychological Foundations of Emotion Persistence and Dynamics}

Psychological research has shown that emotions are not instantaneous responses but dynamic processes that unfold over time \cite{Davidson01051998, verduyn2011relation, bransintensity}. \textit{Emotional inertia} describes the tendency of emotional states to persist and influence next affective experiences rather than changing them instantly \cite{KOVAL20243, liu2024emotionic, waughetal}. Studies show that past emotional states are predictive of future ones and they reflect a form of temporal dependency in affective experience \cite{kuppens2010emotional, koval2015emotional}.

As mentioned before, emotional persistence exists, but it is not infinite. Emotions naturally decay over time and return to a baseline due to context changes and cognitive processes \cite{253d06589a9c483b95609e8bfa2f83e6, verduyn2011relation, bransintensity}. Emotional state durations vary across individuals and situations which indicate that emotions both linger and adapt over time instead of being fixed throughout an entire dialogue \cite{verduyn2011relation, 253d06589a9c483b95609e8bfa2f83e6}. 

In spite of these well-established findings, psychological insights on emotion persistence, decay, and individual differences have not yet been directly incorporated in computational ERC models. Most of the current approaches rely on implicit temporal encoding through recurrent or attention-based architectures, without using emotions as latent states with inertia and speaker-dependent dynamics \cite{Majumder_Poria_Hazarika_Mihalcea_Gelbukh_Cambria_2019, ghosal-etal-2019-dialoguegcn, Ma_2024}.

\section{Methodology}
\label{sec:method}

\subsection{Problem Formulation}

In ERC, the goal is to predict the conveyed emotions for the utterances, using the given conversation up to that point.

A dialogue is represented as an ordered utterance sequence
\begin{equation}
\mathcal{D} = \{ u_1, u_2, \dots, u_T \},
\end{equation}
where $T$ denotes the total number of utterances and the index $i \in \{1, \dots, T\}$ represents the temporal order of the utterances. Each utterance $u_i$ is has a speaker embedding $s_i \in \mathcal{S}$.

Each utterance has three representations in a multimodal setting:
\begin{equation}
u_i = \big( \mathbf{u}_i^{t}, \mathbf{u}_i^{a}, \mathbf{u}_i^{v} \big),
\end{equation}
where $\mathbf{u}_i^{t} \in \mathbb{R}^{D_t}$, $\mathbf{u}_i^{a} \in \mathbb{R}^{D_a}$, and $\mathbf{u}_i^{v} \in \mathbb{R}^{D_v}$ denote the text, acoustic, and visual features of the $i$-th utterance, respectively.

The objective of ERC is to estimate, for every utterance $u_i$, the posterior distribution over emotion classes
\begin{equation}
p(y_i \mid u_1, \dots, u_i),
\end{equation}
where $y_i \in \{1, \dots, C\}$ denotes the emotion label at time step $i$, and $C$ is total number of emotion categories.

\subsection{Multimodal Evidence Encoder with Intra- and Inter-Modal Attention (Baseline)}
\label{subsec:baseline}

\begin{figure*}[ht!]
    \centering
    \includegraphics[
        height=0.49\textheight,
        trim=3cm 4.5cm 0cm 2cm,
        clip
    ]{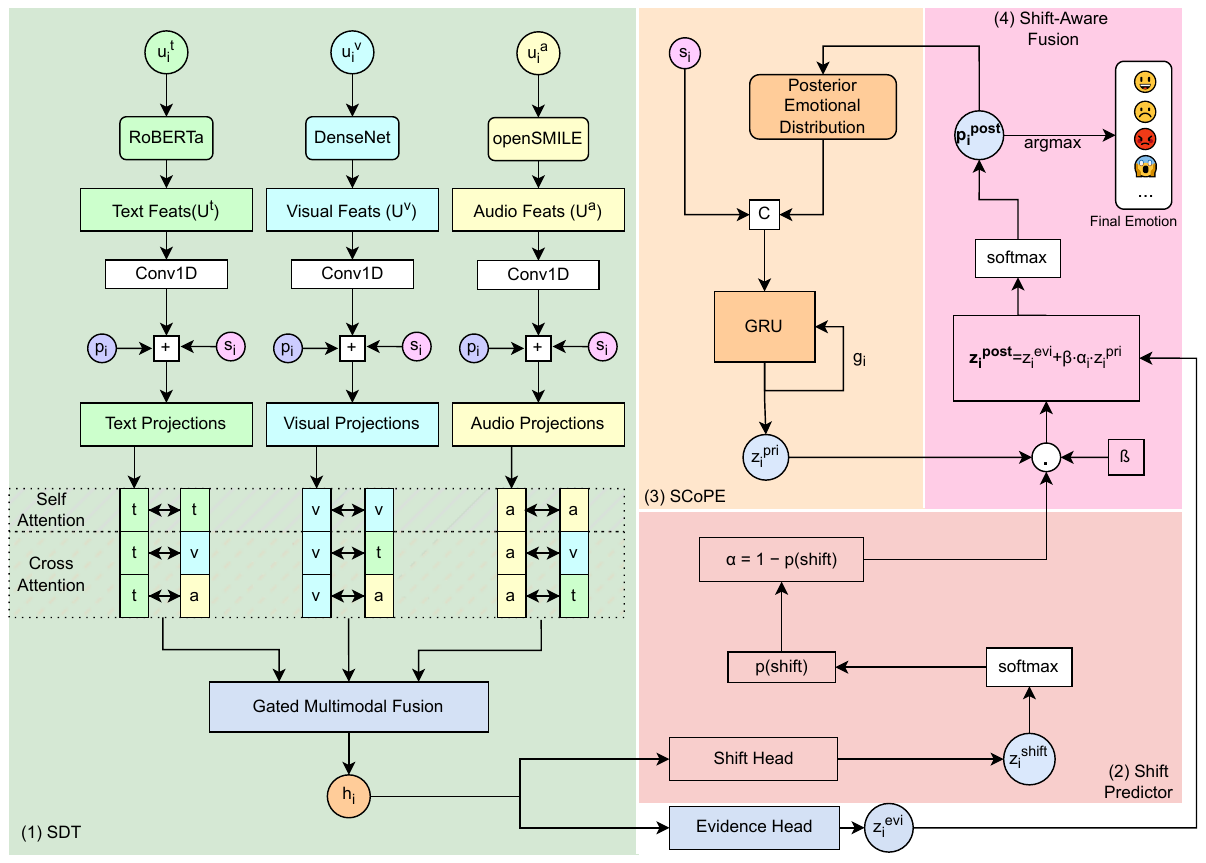}
    \caption{The architecture of our approach, consisting of 4 parts: (1) SDT (Baseline), (2) Shift Predictor, (3) Speaker-Conditioned Priors over Emotions (SCoPE), (4) Emotion Shift-Aware Fusion.}
    \label{fig:architecture}
\end{figure*}

Our multimodal evidence encoder is based on the SDT framework \cite{Ma_2024}, which was a state-of-the-art approach for multimodal ERC on IEMOCAP \cite{busso2008iemocap}. SDT provides a robust utterance-level representation by explicitly modeling intra- and inter-modal interactions. Figure~\ref{fig:architecture}(1) shows the main components of our baseline. We use pre-extracted features for each modality in our input data. Textual features are obtained using RoBERTa-based encoder, visual features are derived from a DenseNet, and acoustic features are extracted using openSMILE. These features are then summed with speaker embeddings and positional embeddings to create modality-wise projections.

First, each modality is processed separately using a modality-specific encoder to obtain utterance-level hidden representations. These encoders aim to extract modality-specific emotional features while preserving the alignment across modalities. Each modality-specific encoder projects the raw features into a shared hidden space:
\begin{equation}
\mathbf{x}_i^m = f_m(\mathbf{u}_i^m), \quad m \in \{t, a, v\},
\end{equation}
where $\mathbf{x}_i^m \in \mathbb{R}^{H}$ denotes utterance-level representation for modality $m$.
 SDT uses intra-modal attention that aggregates token-level or frame-level features into a single vector. Then, inter-modal attention is applied to these separate unimodal representations to further refine them. The intra- and inter-modal attention mechanisms produce modality-aware utterance representations, which are combined with gated multimodal fusion to obtain its final representation:
\begin{equation}
\mathbf{h}_i = \mathrm{Fuse}\big(\mathbf{x}_i^t, \mathbf{x}_i^a, \mathbf{x}_i^v\big),
\end{equation}
where $\mathbf{h}_i \in \mathbb{R}^{H}$ denotes the fused multimodal utterance representation.
 This allows each modality to attend to complementary information from the others. Then, the fused representation is projected into an emotion label space to produce utterance-level emotion logits. These logits represent the model’s multimodal evidence. This approach provides strong multimodal evidence; but it treats each utterance prediction independently. Moreover, it does not explicitly model speaker-specific emotional properties such as continuity or transitions across conversational turns. Finally, the multimodal fused representation is mapped to the emotion label space to produce multimodal evidence logits:
\begin{equation}
\mathbf{z}_i^{evi} = W_e \mathbf{h}_i,
\end{equation}
where $\mathbf{z}_i^{evi} \in \mathbb{R}^{C}$ represents the evidence scores over $C$ emotion classes.

\subsection{Emotion Shift Prediction}

As mentioned previously, even though emotions tend to persist across multiple turns across dialogues, sudden changes may happen due to new context or interaction factors. To cover such cases, we introduce an \textit{emotion shift prediction} module that estimates whether the emotional state changes between consecutive utterances. The operations in the emotional shift head can be seen in Figure~\ref{fig:architecture}(2).

Using the fused multimodal utterance representation $\mathbf{h}_i$ (Section~\ref{subsec:baseline}), we predict shift logits using a separate classification head:
\begin{equation}
\mathbf{z}_i^{\text{shift}} = W_s \mathbf{h}_i,
\end{equation}
where $\mathbf{z}_i^{\text{shift}} \in \mathbb{R}^{2}$ corresponds to the scores for \textit{no-shift} and \textit{shift}. The probability of an emotional shift at utterance $i$ is obtained as:
\begin{equation}
p_{\text{shift}}(i) = \mathrm{softmax}(\mathbf{z}_i^{\text{shift}}).
\end{equation}

In the later fusion module, the obtained estimated shift probability is used as a confidence signal to control the influence of temporal emotional memory, rather than a direct prediction target. We define a memory confidence weight as:
\begin{equation}
\alpha_i = 1 - p_{\text{shift}}(i),
\end{equation}
so higher likelihood of emotional shift prediction reduces the reliance on the emotional memory. 

Although both the multimodal evidence logits $\mathbf{z}_i^{evi}$ and the shift logits $\mathbf{z}_i^{\text{shift}}$ are computed from the same fused utterance representation $\mathbf{h}_i$, they correspond to two distinct prediction heads with different objectives and supervision signals. The evidence head is trained to predict emotion categories, whereas the shift head is trained using an auxiliary supervision signal that shows whether an emotional change occurs between consecutive utterances. This separation allows the shift predictor to model temporal discontinuities independently of emotion classification, while still influencing the final prediction through shift-aware fusion.

\subsection{Speaker-Conditioned Priors over Emotions (SCoPE)}

Multimodal evidence provides strong utterance-level cues, but it does not explicitly encode temporal dependencies or speaker-conditioned dynamics of emotions. To address this limitation, we introduce a \textit{Speaker-Conditioned Priors over Emotions (SCoPE)}, which models emotional state evolutions throughout the conversations.

We implement this prior as an autoregressive GRU module that maintains a latent space capturing temporal emotional patterns, in other words, emotional memory. It is important to highlight that SCoPE does not see the current utterance evidence; instead, it is conditioned solely on the posterior belief from the previous time step and the current speaker identity. An overview of SCoPE is shown in Figure~\ref{fig:architecture}(3). The modeling of SCoPE helps to separate the evidence modeling and temporal prior estimation.  

At the beginning of a dialogue, the posterior belief is initialized to a uniform distribution:
\begin{equation}
\mathbf{p}_0^{\text{post}} = \frac{1}{C}\mathbf{1},
\end{equation}
where $C$ denotes the number of emotion classes.

At each utterance $i$, the GRU updates its hidden state according to:
\begin{equation}
\mathbf{g}_i = \mathrm{GRU}\!\left(
\mathbf{g}_{i-1},
\left[\mathbf{p}_{i-1}^{\text{post}} \;\Vert\; \mathbf{s}_i\right]
\right),
\end{equation}
where $\mathbf{s}_i$ denotes the embedding of the active speaker at time step $i$, and $\Vert$ represents vector concatenation. The updated hidden state is then projected to produce prior emotion logits:
\begin{equation}
\mathbf{z}_i^{pri} = W_{pri} \mathbf{g}_i,
\end{equation}
where $\mathbf{z}_i^{pri} \in \mathbb{R}^{C}$ represents the speaker-conditioned prior over emotion classes.

This representation allows our model to capture speaker-dependent emotional persistence and transition patterns without direct access to the current multimodal evidence.

\subsection{Shift-Aware Fusion}

The multimodal evidence encoder and the speaker-conditioned emotional prior provide complementary information: the former captures the current emotional cues, while the latter encodes temporal continuity. To balance these two sources dynamically, we propose a \textit{shift-aware fusion} mechanism guided by the predicted emotion shift probability (Figure~\ref{fig:architecture}(4)). The posterior emotion logits are computed by combining the multimodal evidence logits $\mathbf{z}_i^{evi}$ (Section~3.2) and the prior logits $\mathbf{z}_i^{pri}$ as follows:
\begin{equation}
\mathbf{z}_i^{\text{post}} =
\mathbf{z}_i^{evi}
+
\beta \cdot \alpha_i \cdot \mathbf{z}_i^{pri},
\end{equation}
where $\alpha_i = 1 - p_{\text{shift}}(i)$ controls the contribution of the prior, and $\beta$ is a scalar that modulates the overall influence of temporal memory.

The final posterior distribution over emotion classes is obtained via softmax normalization:
\begin{equation}
\mathbf{p}_i^{\text{post}} = \mathrm{softmax}(\mathbf{z}_i^{\text{post}}).
\end{equation}

At inference time, the predicted emotion label for utterance $i$ is selected using a maximum of a posterior decision:
\begin{equation}
\hat{y}_i = \arg\max_{c} \mathbf{p}_i^{\text{post}}[c].
\end{equation}

This shift-aware fusion strategy enables the model to rely more on temporal emotional context when emotional continuity is likely. On the other hand, it allows for a quick adaptation to sudden emotional changes when shifts are detected.

\section{Experimental Setup}
\label{sec:experiments}

\subsection{Datasets}

We use two ERC datasets for this work: IEMOCAP and MELD. They are well known benchmark ERC datasets comprising text, visual, and audio modalities. Since we are using SDT as a baseline, we decided to use the feature representations released by the SDT repository for fair comparison and consistency.

IEMOCAP is an ERC dataset containing scripted dialogues that has 5 sessions, including 151 long dialogues with approximately 7430 utterances. It has ten emotion labels: Neutral, Happiness, Sadness, Anger, Excited, Frustration, Fear, Surprise, Disgust, and Other (Uninformative). Disgust does not appear in the validation split. Also, Fear and Surprise are quite rare, so they are excluded from many studies. Commonly, IEMOCAP tasks are: 8-way (without Disgust and Other), 6-way (without Surprise and Fear), or 4-way (Neutral, Happiness, Sadness and Anger). We use the 6-way classification in this work, which is the most commonly used task for this dataset in the literature.

MELD is a dataset gathered from scenes of the ``Friends" TV show. MELD is a much more challenging dataset than IEMOCAP because the data has high noise, imbalance, and contextual bias since it is a comedy show. It has approximately 1400 dialogues with 13000 utterances with 7 emotion labels: Neutral, Joy, Sadness, Anger, Surprise, Disgust, and Fear. Neutral is the most common label among all. Fear and Disgust are quite rare regarding others. 

\subsection{Training Configuration}

We trained each of our models for 150 epochs. Table~\ref{tab:hyperparams} shows our hyperparameters used for reproducibility purposes. We used the Optuna\footnote{\url{https://optuna.org/}} hyperparameter optimization tool to sweep over those parameters to have our final values. For all of the training, our architecture required approximately 3 GB of VRAM of hardware. For experiments, we trained 10 models for each of those datasets and report mean and standard deviation for reliable results.

\begin{table}[!htbp]
\centering
\caption{Hyperparameters used for each dataset.}
\label{tab:hyperparams}
\begin{tabular}{lcc}
\hline
\textbf{Hyperparams} & \textbf{IEMOCAP} & \textbf{MELD} \\
\hline
batch\_size      & 16      & 8      \\
learning rate           & $1e{-4}$ & $5e{-6}$ \\
dropout      & 0.5     & 0.5     \\
num\_layers  & 1       & 1       \\
ß  & 1.5    & 1    \\
\hline
\end{tabular}
\end{table}

\subsection{Loss Functions}

We trained our models end-to-end using a multi-task objective. 
The primary training signal is cross-entropy loss applied to the final posterior emotion distribution, which supervises the overall emotion prediction.

To train the emotion shift prediction head, we extract an additional supervision signal from the datasets, which we refer to as \textit{Emotion Shift}. This is an utterance level signal which takes a binary value of 1 if the emotion label changes between two consecutive lines of the same speaker, and 0 otherwise. The extracted shift labels are used to supervise the shift logits $\mathbf{z}_i^{\text{shift}}$ via a cross-entropy loss with a loss weight of 0.3. This objective enables the model to explicitly learn when emotional continuity is likely to be disrupted.

For the SCoPE, we introduce an additional prior supervision loss. This loss encourages the prior logits $\mathbf{z}_i^{pri}$ to align with the ground-truth emotion labels, while being weighted with the predicted likelihood of emotional continuity. Specifically, the prior loss is defined as:
\begin{equation}
\mathcal{L}_{\text{prior}} = - \sum_{i=1}^{T} \big(1 - p_{\text{shift}}(i)\big)\log p_{\text{prior}}(y_i),
\end{equation}
and is applied with a loss weight of 0.2.

Additionally, the baseline SDT~\cite{Ma_2024} framework has auxiliary unimodal cross-entropy losses to the predictions from individual modalities in order to stabilize multimodal learning. Furthermore, an auxiliary knowledge distillation loss is employed between unimodal predictions and the multimodal evidence prediction to encourage the model be consistent across modalities.

Total training loss is computed as a weighted combination of the posterior emotion loss, the prior supervision loss, and the auxiliary objectives.

\subsection{Evaluation Metrics}

We obtained W-F1 and Accuracy metrics for IEMOCAP and MELD, as they are the most common metrics used with those datasets. We present the average over ten models we trained for both metrics, as well as the best-performing one among them. In the literature, models achieve higher scores on IEMOCAP than on MELD in both metrics, showing that MELD is a much more complex dataset. Additionally, we gathered these scores for each emotion separately to show the most affected emotions by our approach. 

\begin{table*}[!htp]
\centering
\caption{Performance comparison of our SCoPE architecture on the IEMOCAP dataset. Emotion-wise F1 scores are reported for each category, along with overall Accuracy (Acc) and Weighted F1 (WF1). The first row of the SCoPE shows the average results of ten random runs and the second one ($\dagger$) indicates the best one among them. Best results in each column are highlighted in bold.}
\label{tab:iemocap_results}
\resizebox{\textwidth}{!}{%
\begin{tabular}{l c c c c c c c c c}
\toprule
\textbf{Method} & \textbf{Year} & \textbf{Happy} & \textbf{Sad} & \textbf{Neutral} & \textbf{Angry} & \textbf{Excited} & \textbf{Frustrated} & \textbf{Acc} & \textbf{WF1} \\
\midrule

JOYFUL\cite{li-etal-2023-joyful}     & 2023 & 60.94 & 84.42 & 68.24 & 69.95 & 73.54 & 67.55 & 70.55 & 71.03 \\
AdaIGN\cite{Tu_Xie_Liang_Wang_Xu_2024}     & 2024 & 53.04 & 81.47 & 71.26 & 65.87 & 76.34 & 67.79 & 70.49 & 70.74 \\
CFN-ESA\cite{li2024cfn}    & 2024 & 56.76 & 81.34 & 71.19 & 68.23 & 75.83 & 65.50 & 70.69 & 70.61 \\

SDT (Baseline)~\cite{Ma_2024}        & 2024 & 64.04 & 80.39 & 72.88 & 69.77 & 77.88 & 68.61 & 72.57 & 72.02 \\
DER-GCN\cite{10458270}    & 2025 & 58.80 & 79.80 & 61.50 & \textbf{72.10} & 73.30 & 67.80 & 69.70 & 69.40 \\

GS-MCC\cite{Ai_Zhang_Shou_Meng_Chen_Li_2025}     & 2025 & 65.40 & 81.20 & 70.90 & 70.80 & 81.40 & 71.00 & 73.80 & 73.90 \\
SEDC\cite{10900452}       & 2025 & 62.78 & \textbf{84.49} & 71.52 & 65.71 & 73.63 & 68.09 & 71.60 & 71.68 \\
PCDS\cite{pcds}       & 2025 & 54.74 & 78.62 & 71.78 & 69.79 & 74.74 & 68.53 & 70.84 & 70.89 \\

\midrule
\multirow{2}{*}{\textbf{SCoPE}} & \multirow{2}{*}{2026} 
& \textbf{66.00$\pm$1.0} & 83.00$\pm$1.1 & \textbf{73.90$\pm$1.0} & 70.60$\pm$0.8 & \textbf{82.00$\pm$1.0} & \textbf{69.50$\pm$1.0} & \textbf{74.57$\pm$0.4} & \textbf{74.68$\pm$0.4} \\

& 
& 66.90$^{\dagger}$ & 82.70$^{\dagger}$ & 75.10$^{\dagger}$ & 70.80$^{\dagger}$ 
& 84.00$^{\dagger}$ & 71.30$^{\dagger}$ & 75.72$^{\dagger}$ & 75.82$^{\dagger}$ \\
\bottomrule
\end{tabular}}
\end{table*}

\subsection{Baseline Models}

In Section~\ref{sec:results}, we compare the performance of our proposed model with eight recent state-of-the-art methods that report results using the same evaluation metrics on the same benchmark datasets. These models were selected to provide a comprehensive and up-to-date comparison on the multimodal ERC task. Based on their architectural characteristics, these models can be grouped as graph-based~\cite{Tu_Xie_Liang_Wang_Xu_2024, li-etal-2023-joyful, 10458270}, transformer-based~\cite{li2024cfn, Ma_2024}, contrastive learning-based~\cite{Ai_Zhang_Shou_Meng_Chen_Li_2025, 10900452, li-etal-2023-joyful} and auxilary task learning~\cite{pcds, li2024cfn, Ma_2024} models.

\section{Results \& Analyses}
\label{sec:results}
Table~\ref{tab:iemocap_results} shows our results for IEMOCAP, and Table~\ref{tab:meld_results} shows the results for the MELD dataset for W-F1 and the Accuracy scores of overall and emotion-wise classification. Additionally, the results from our baseline model (SDT) and eight recent state-of-the-art methods have been added to the tables for comparison.

\begin{table*}[!htbp]
\centering
\caption{Performance comparison on the MELD dataset. Metric and display style are as in Table~\ref{tab:iemocap_results}.}
\label{tab:meld_results}
\resizebox{\textwidth}{!}{%
\begin{tabular}{l c c c c c c c c c c}
\toprule
\textbf{Name} & \textbf{Year} & \textbf{Neutral} & \textbf{Surprise} & \textbf{Fear} & \textbf{Sadness} & \textbf{Joy} & \textbf{Disgust} & \textbf{Anger} & \textbf{Acc} & \textbf{WF1} \\
\midrule
JOYFUL\cite{li-etal-2023-joyful} & 2023 & 76.80 & 51.91 & -- & 41.78 & 56.89 & -- & 50.71 & 62.53 & 61.77 \\
AdaIGN\cite{Tu_Xie_Liang_Wang_Xu_2024} & 2024 & 79.75 & \textbf{60.53} & 15.20 & \textbf{43.70} & 64.54 & 29.30 & 56.15 & 67.62 & 66.79 \\
CFN-ESA\cite{li2024cfn} & 2024 & 79.93 & 58.47 & 22.41 & 41.16 & 64.78 & 30.14 & 53.91 & 66.36 & 67.42 \\
SDT (Baseline)~\cite{Ma_2024} & 2024 & 75.08 & 56.20 & 22.08 & 41.42 & 63.14 & 31.18 & 50.28 & 62.01 & 63.38 \\
DER-GCN\cite{10458270} & 2025 & 80.60 & 51.00 & 10.40 & 41.50 & 64.30 & 10.30 & \textbf{57.40} & 66.80 & 66.10 \\
GS-MCC\cite{Ai_Zhang_Shou_Meng_Chen_Li_2025} & 2025 & \textbf{81.80} & 58.30 & 23.80 & 35.80 & \textbf{66.40} & 30.70 & 54.40 & \textbf{68.10} & \textbf{69.00} \\
SEDC\cite{10900452} & 2025 & 80.09 & 58.11 & 23.38 & 40.62 & 65.22 & 28.30 & 52.16 & 67.43 & 66.16 \\
PCDS\cite{pcds} & 2025 & 79.03 & 56.79 & -- & 32.66 & 57.07 & -- & 48.67 & 64.33 & 62.61 \\

\midrule
\multirow{2}{*}{\textbf{SCoPE}} & \multirow{2}{*}{2026}
& 79.50$\pm$0.4 & 57.50$\pm$0.7 & \textbf{26.00$\pm$2.4} & 41.20$\pm$1.0 & 63.00$\pm$0.5 & \textbf{32.20$\pm$1.3} & 52.00$\pm$1.0 & 65.76$\pm$0.2 & 65.65$\pm$0.1 \\

& 
& 80.40$^{\dagger}$ & 56.90$^{\dagger}$ & 26.80$^{\dagger}$ & 43.10$^{\dagger}$
& 62.90$^{\dagger}$ & 32.10$^{\dagger}$ & 50.80$^{\dagger}$ & 66.25$^{\dagger}$ & 66.01$^{\dagger}$ \\

\bottomrule
\end{tabular}}
\end{table*}

\subsection{Overall Classification Results}

For the IEMOCAP dataset, SCoPE achieved 75.72\% Accuracy and 75,82\% in WF1. Averaged over 10 random runs, SCoPE achieved 74.57\% Accuracy and 74.68\% in WF1. Both results demonstrate SCoPE's superior performance over recent state-of-the-art models on the IEMOCAP dataset in multimodal settings. In Emotion-wise scores, SCoPE performs best in Happy and Neutral, while still being competitive in other emotions too. 
For MELD, it does not perform as the best model, however, the baseline also has lower performance on this dataset. This could be because MELD is more noisy, in terms of having more exaggerated context (coming from it being a comedy show), more background audio, and many more speakers.  Still, SCoPE increased the scores over its baseline and it managed to achieve the best results in Fear and Disgust labels among all models, which are the rarest labels in MELD. Also, given that Happy has the lowest label count in IEMOCAP as well, it can be concluded that our work enables the model to benefit from the rare labels more, while remaining competitive in other labels. 

\begin{figure*}[t!]
    \centering
    \includegraphics[width=1\textwidth]{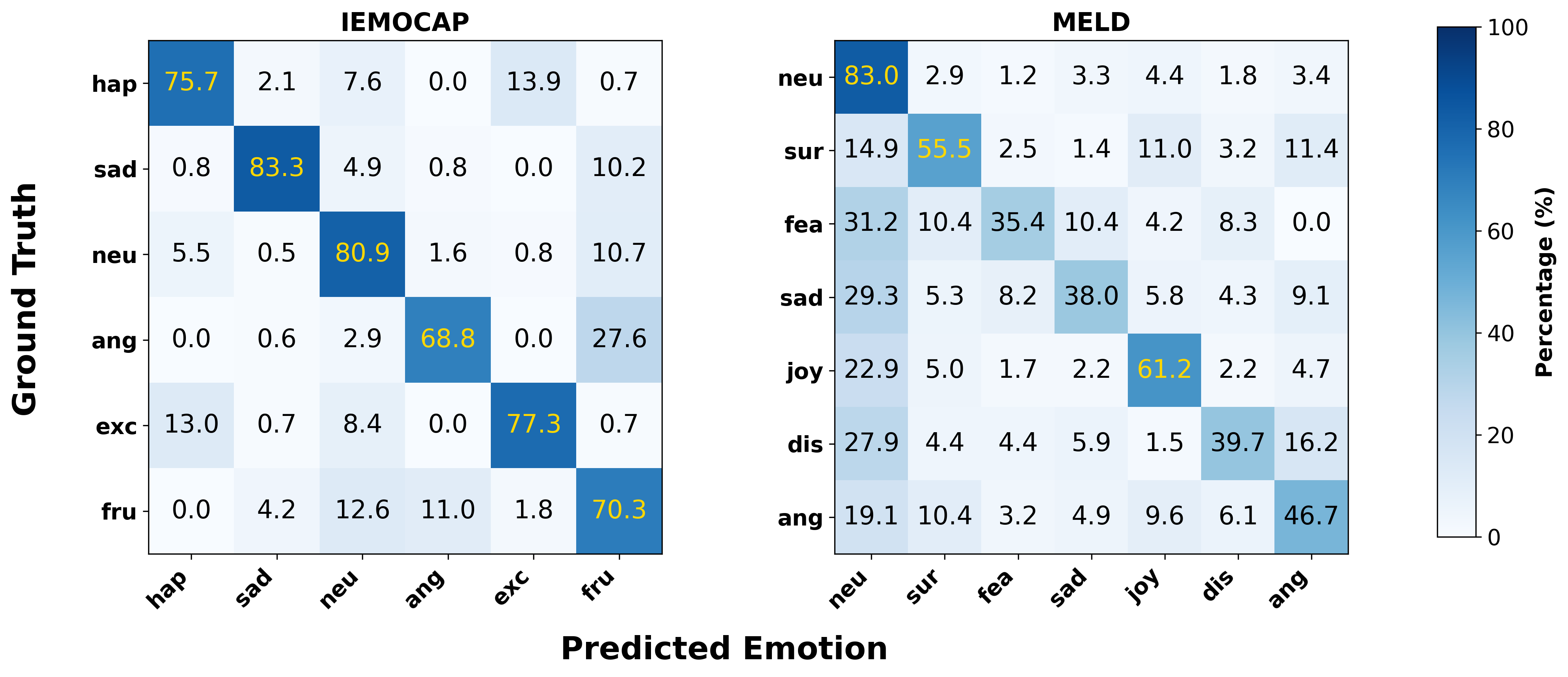}
    \caption{\centering Confusion matrices of our best performing model, for IEMOCAP and MELD datasets respectively. The classified quantities are converted into percentages for displaying purposes.}
    \label{fig:confusion}
\end{figure*}

\begin{figure}[!t]
\centering
\includegraphics[width=0.495\linewidth,clip,trim={0.2cm 0 0.2cm 0.8cm}]{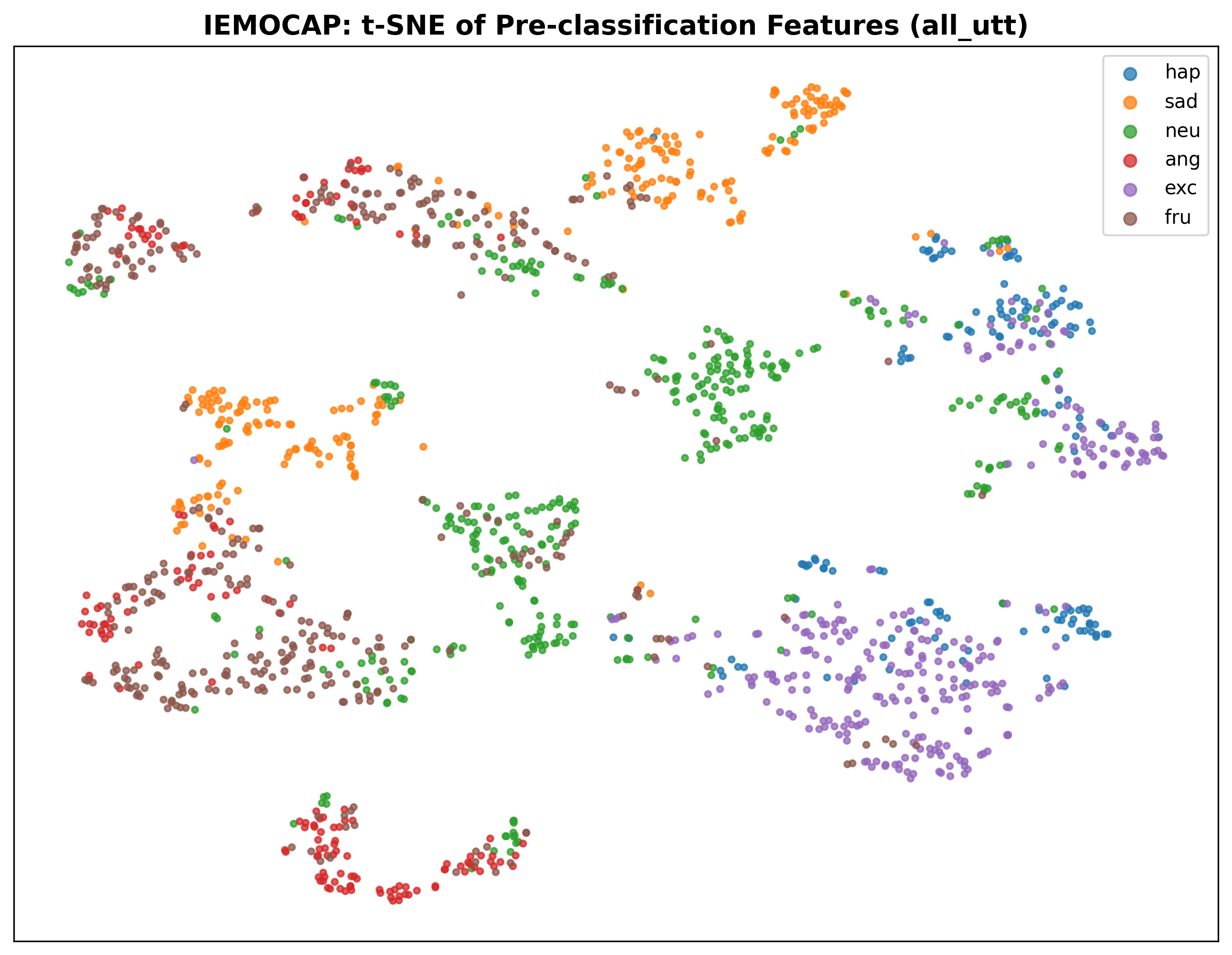}
\includegraphics[width=0.495\linewidth,clip,trim={0.2cm 0 0.2cm 0.8cm}]{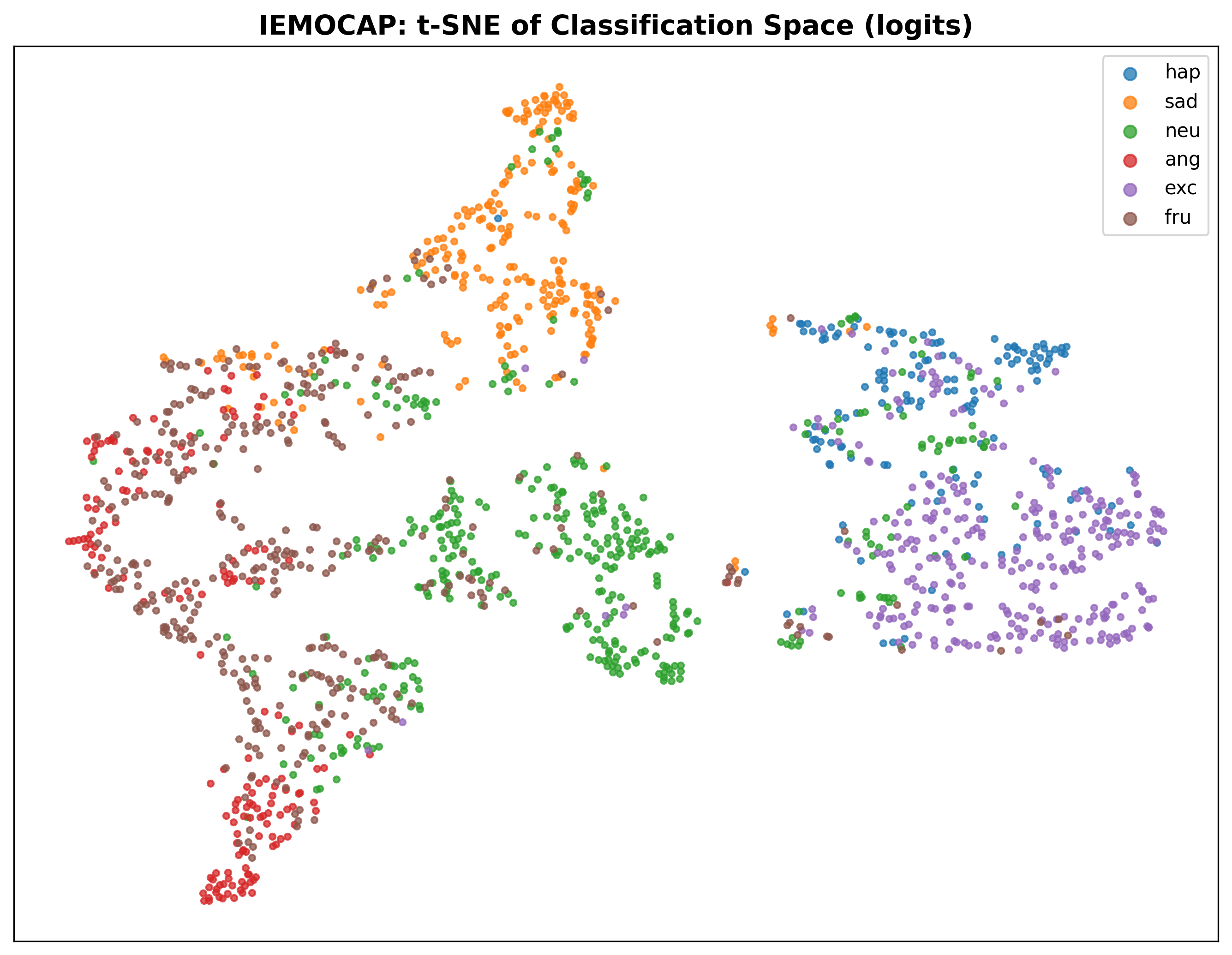}

\caption{t-SNE visualizations of IEMOCAP features before classification (left) and after classification (right).}
\label{fig:iemocap_tsne}
\end{figure}

\begin{figure}[!b]
\centering
\includegraphics[width=0.495\linewidth,clip,trim={0.2cm 0 0.2cm 0.8cm}]{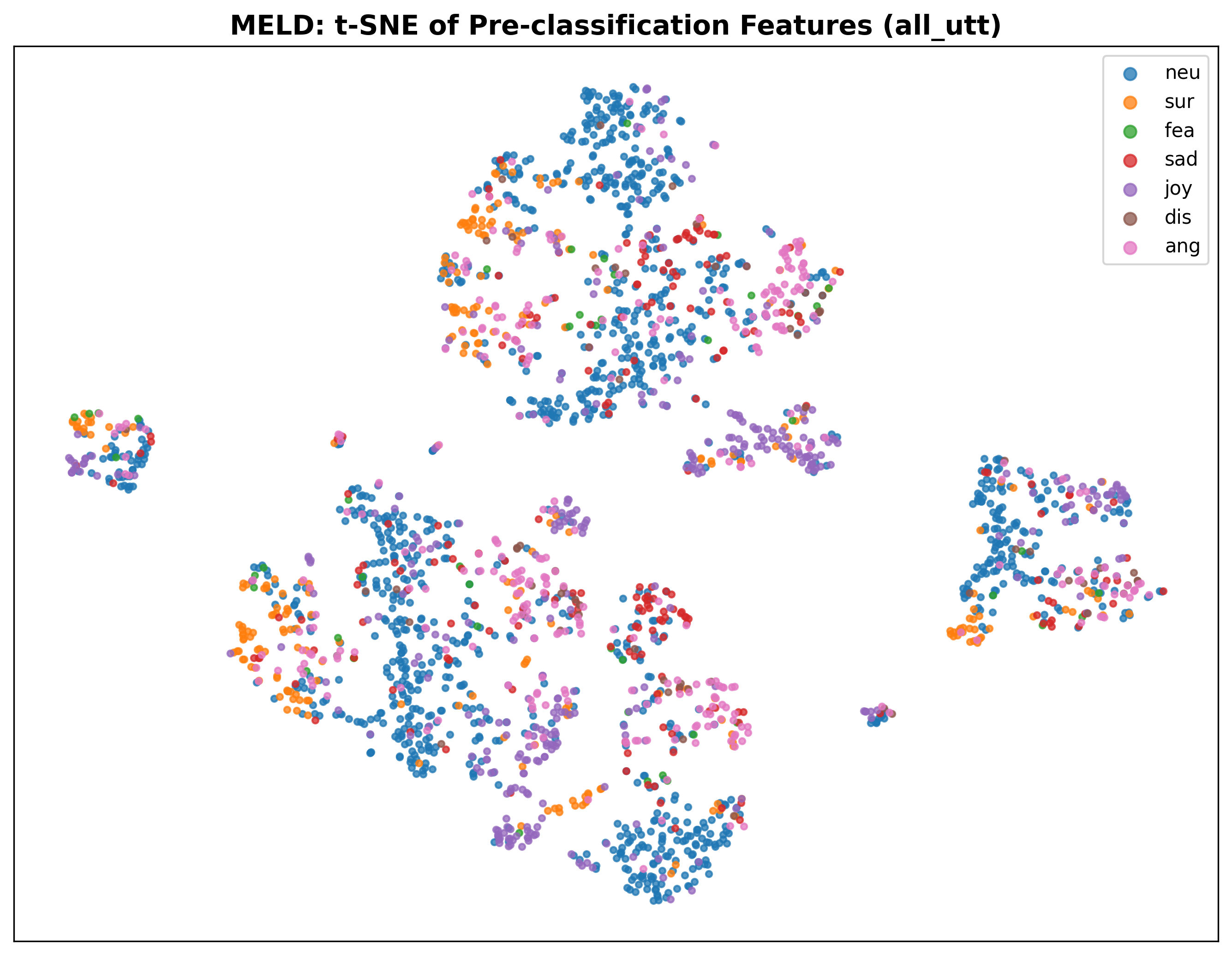}
\includegraphics[width=0.495\linewidth,clip,trim={0.2cm 0 0.2cm 0.8cm}]{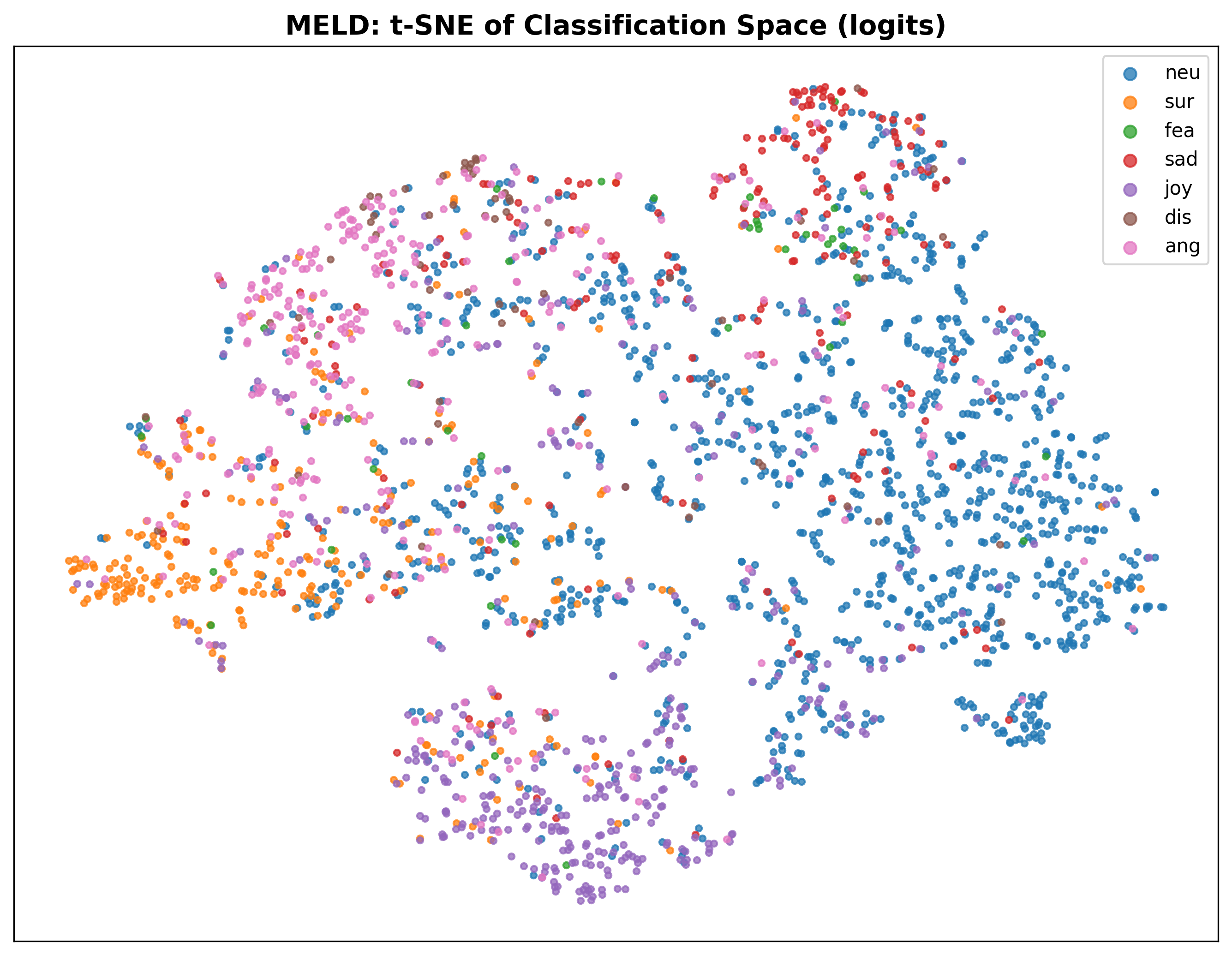}

\caption{t-SNE visualizations of MELD features before classification (left) and after classification (right).}
\label{fig:meld_tsne}
\end{figure}

Figure~\ref{fig:confusion} shows the confusion matrices of SCoPE for both datasets. Especially in MELD, there are still lots of samples wrongly predicted as Neutral. This is expected because MELD has an extreme imbalance towards the Neutral emotion. However, it is still noticable that our model benefits from Fear, Sadness, and Disgust being much lower in numbers. For IEMOCAP, the matrix shows more correct classifications because that dataset provides much more balanced and cleaner data. Figure~\ref{fig:iemocap_tsne} and Figure~\ref{fig:meld_tsne} show how the multimodal features of IEMOCAP and MELD distribute in feature space and how our models cluster them after the training. First, it can be noticed that the model struggles more to distinguish Anger-Frustration, Happiness-Excited duos on IEMOCAP and Sadness-Fear and Neutral against every other label for MELD. This shows the high correlation of those emotion duos semantically, causing the model to struggle more on those emotion duos than other emotions. For the case of the emotion Neutral for MELD, the imbalance problem is shown clearly in visual. Still, the labels are clustered well enough by the model visually in both datasets, showing its competitive performance, even for rare labels.

Finally, we compare SCoPE against its baseline (SDT) in Table~\ref{tab:iemocap_results} and Table~\ref{tab:meld_results}). It can be seen that the baseline gets a consistent performance boost across most of the emotions, as well as the overall scores in both datasets. That shows that performance improvements can be achieved with lightweight additions such as GRUs. They harmonize well with ERC models because of their autoregressive nature. Since emotional persistance in dialogues is typically short-to-mid range, it matches perfectly with GRU's single gated state. So the module still has enough memory to smooth the emotions, but not unnecessarily long to hallucinate. Combined with the shift prediction head and shift-aware fusion, our modules complement the model's performance effectively.

\subsection{Ablation Study: Effects of Different ß values}

\begin{figure}[htbp!]
    \centering
    \includegraphics[
        width=0.5\textwidth,
        trim=0 0 0 0cm,
        clip
    ]{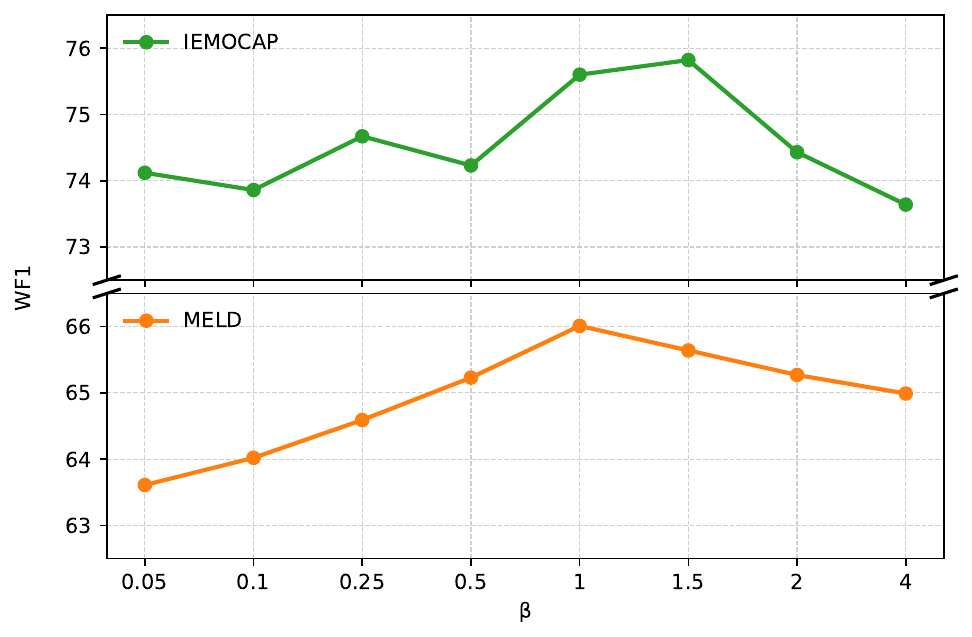}
    \caption{\centering The WF1 scores over different $\beta$ values.}
    \label{fig:beta}
\end{figure}

We performed an additional ablation study in which we tested how different scaling factors (ß) of the shift-aware fusion affect the model performance. We conducted multiple train runs with different ß values, and saved the best models across the runs. Figure~\ref{fig:beta} shows the results of those experiments for both datasets. Low values like 0.05 and 0.1 would make the model more sensitive to immediate emotion shifts, whereas high values like 4 would make the model more conservative and faithful to the prior.

From the Figure~\ref{fig:beta} it can be seen that the best ß values for both datasets are between 1 and 2. Having lower scores on extreme ß values is expected, and our shift-aware fusion module tries to find an optimal balance between the prior and the evidence with this parameter. Additionally, optimal intervals for each dataset show the characteristics of those datasets and how sensitive the balance should be. The optimal ß for MELD is lower than for IEMOCAP, showing that it is more prone to having more sudden or abrupt emotional changes during its conversations. Given that MELD data is taken from a comedy TV show, it is reasonable that MELD has a lower optimal ß than IEMOCAP, which is recorded with more stable scripts in an isolated environment. Moreover, MELD has more speakers than IEMOCAP, which is also a possible contributing factor.

\subsection{Inference Time}

In practice, our architecture requires only 3.8 GB of GPU memory (VRAM) during inference, which makes it feasible to run on consumer-level GPUs without specialized hardware requirements. To also measure its potential for real-time deployment, we further evaluate the inference efficiency of our model by measuring both per-utterance latency and total evaluation time on the benchmark dataset's extracted features. On the MELD test set, our model takes an average inference time of 0.25 ms per utterance, with a total evaluation time of 0.66s for the entire test split. Similarly, on the IEMOCAP dataset, the model requires 0.20 ms per utterance on average and completes full test-set evaluation in 0.33 s.

These results demonstrate that the proposed approach is more computationally efficient than most of the current models. Given that a lot of models nowadays are either based on or tied to LLM's, the need of smaller models is increasing for user-level applications, or based on reduction of carbon footprint. Thanks to its low latency, small memory allocation, and reliance on lightweight recurrent and fusion modules, the model is well-suited for real-time emotion recognition applications, such as socially interactive robotics, human–computer interaction systems, and on-device affective computing.

\section{Conclusion}
\label{sec:conc}
As shown in prior psychological research, emotional persistence and inertia are part of human cognition. Therefore, related tasks that are close to these concepts semantically, such as ERC, should not be treated as isolated utterance classification. Instead, they should track and update the evolving affective states of the speakers. In this paper, we introduced SCoPE, a module that models human implicit emotional priors. Second, we integrated emotion shift prediction, a widely studied method in ERC, as a control signal for our system. Finally, we combined SCoPE and emotion shift with shift-aware fusion to balance the speaker history versus the evidence. Experimental results demonstrate that this approach consistently improves the performance over its baseline, showing that temporal modeling is useful and provides complementary information when it is selectively applied. Additionally, we demonstrated that human priors are useful, especially for rare labels that challenge existing ERC models most. Finally, our model outperforms recent models on IEMOCAP while still being very lightweight and fast, making it a very suitable option for real time applications.

SCoPE assumes smooth evolution while it may struggle with sudden changes. Therefore, our future work will model longer-term user behavior beyond individual dialogues to integrate more long term properties of human cognition, such as persona or longer-term mood states of a person. Because priors are a contributing factor, but not the only factor. Additionally, it would be interesting to extend the priors to continuous state modeling with valence/arousal values. Finally, given its applicability, SCoPE as a plug-in for LLM-based conversational agents can be a useful tool to be used in social robotics settings to give robotic agent better knowledge about the user emotions.

\section*{Declarations}

\subsection*{Acknowledgments}

The authors gratefully acknowledges the financial support provided by Horizon Europe under the MSCA grant agreement No 101168792 (SWEET) and No 101072488 (TRAIL), and the Ministry of National Education of the Republic of Türkiye (YLSY Scholarship Program).

\subsection*{Data Availability}

The authors declare that this manuscript does not have data generation or analysis.

\subsection*{Funding}
This work was supported by Horizon Europe under the MSCA grant agreement No 101168792 (SWEET) and No 101072488 (TRAIL), and Ministry of National Education of the Republic of Türkiye.

\subsection*{Competing Interests}

The authors declare no competing interests.

\bibliography{custom4}


\end{document}